\documentclass[10pt,twocolumn,letterpaper]{article}

\usepackage{iccv}
\usepackage{times}
\usepackage{epsfig}
\usepackage{graphicx}
\usepackage{amsmath}
\usepackage{amssymb}
\usepackage{rotating}
\usepackage{tablefootnote}
\usepackage{bbm}
\usepackage{float}
\usepackage[caption = false]{subfig}
 \usepackage{array,multirow,graphicx}

\usepackage[pagebackref=true,breaklinks=true,letterpaper=true,colorlinks,bookmarks=false]{hyperref}

\iccvfinalcopy 

\def\iccvPaperID{23} 

\ificcvfinal\fi
\begin{document}
\newcommand{\JM}[1]{\textcolor{red}{ }}

\title{VACL: Variance-Aware Cross-Layer Regularization for Pruning Deep Residual Networks}


\author{\hspace{-0.15cm} Shuang Gao \\ NVIDIA \and \hspace{-0.25cm} Xin Liu \\ NVIDIA \and \hspace{-0.25cm} Lung-Sheng Chien \\ NVIDIA \and \hspace{-0.25cm}William Zhang \\ NVIDIA \and \hspace{-0.25cm} Jose M. Alvarez \\ NVIDIA}


\maketitle

\begin{abstract}
Improving weight sparsity is a common strategy for producing light-weight deep neural networks. 
However, pruning models with residual learning is more challenging. 
In this paper, we introduce Variance-Aware Cross-Layer (VACL), a novel approach to address this problem. VACL consists of two parts, a Cross-Layer grouping and a Variance Aware regularization. In Cross-Layer grouping the $i^{th}$ filters of layers connected by skip-connections are grouped into one regularization group. Then, the Variance-Aware regularization term takes into account both the first and second-order statistics of the connected layers to constrain the variance within a group.
Our approach can effectively improve the structural sparsity of residual models. For CIFAR10, the proposed method reduces a ResNet model by up to 79.5\% with no accuracy drop, and reduces a ResNeXt model by up to 82\% with $<1\%$ accuracy drop. For ImageNet, it yields a pruned ratio of up to $63.3\%$ with $<1\%$ top-5 accuracy drop. Our experimental results show that the proposed approach significantly outperforms other state-of-the-art methods in terms of overall model size and accuracy. 
   
\end{abstract}

\section{Introduction}
Deep neural networks have shown great success in various applications. However, they are also well known for their heavy computation and storage cost. Numerous efforts have been made to tackle this problem \cite{NIPS2016_6372,NIPS2017_Decomp,NIPS2015_5647,NIPS2014_5544,Dong_2017_CVPR,Gordon_2018_CVPR,1510.00149,1704.04861,1602.07360,BMVC.28.88, Liu_2015_CVPR,Sandler_2018_CVPR,Zhang_2018_CVPR,Tung_2018_CVPR}. One popular strategy is structural model pruning \cite{NIPS2016_6372,  He_2017_ICCV,1608.08710,Luo_2017_ICCV,Scardapane:2017:GSR:3067301.3067328,NIPS2016_6504}. It removes groups of insignificant parameters from the original model based on importance metrics so that the pruned model has fewer parameters with negligible loss of accuracy.

\begin{figure}[t]
\begin{center}
\begin{tabular}{ccc}
\includegraphics[width=0.2\linewidth]{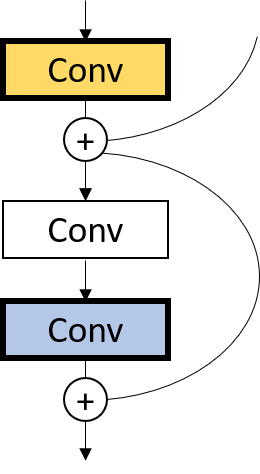}&
\includegraphics[width=0.32\linewidth]{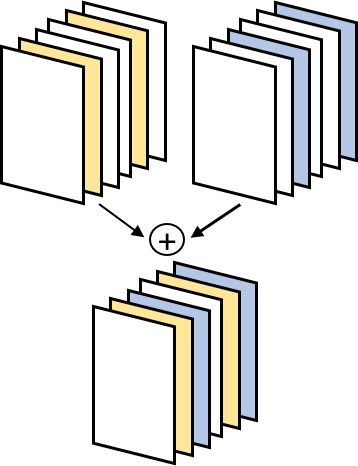}&
\includegraphics[width=0.32\linewidth]{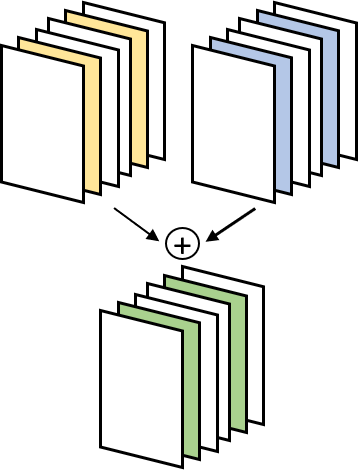}\\
(a)&(b)&(c)\\
\end{tabular}
\end{center}
   \caption{Pruning residual layers (a) is a challenging task. b) Applying sparsity constraints individually to each layer leads to different sparsity structures and therefore more channels need to be retained (colored blocks). c) Aligned sparsity between layers tied by skip connections would help on effective pruning. In this case, only two channels would remain after the pruning operation.}
\label{fig:motivation}
\vspace{-0.5cm}
\end{figure}

Residual learning, initially introduced with ResNet models applied to various vision tasks~\cite{He_2016_CVPR}, has now become a standard component in the design of modern network architectures such as ResNet-v2 \cite{resnet_v2}, Wide ResNet \cite{wresnet}, ResNeXt \cite{Xie_2017_CVPR}, Inception-ResNet \cite{inception-resnet}, Xception \cite{1610.02357} and MobileNet-v2 \cite{Sandler_2018_CVPR}. The key idea behind residual learning is the use of skip-connections and element-wise addition as shown in Fig.~\ref{fig:motivation}{\color{red}.a}. 

Pruning architectures with skip-connections and element-wise additions has two main challenges. First, pruning methods need to obtain sparsity patterns that are aligned between layers connected by element-wise additions (Fig.~\ref{fig:motivation}{\color{red}.c}). The lack of alignment would lead to layers with different sparsity structures and thus, would reduce the pruning effect (see Fig.~\ref{fig:motivation}{\color{red}.b}). Second, using connected layers increases the number of parameters that need to be zeroed simultaneously. As a consequence, the variance of these parameters increases and it becomes more difficult to set them all to zero. Existing solutions to the first problem include pruning only layers that are not connected by skip-connections~\cite{Luo_2017_ICCV}, applying the sparsity constraints only to the identity path~\cite{1608.08710} or using mixed block connectivity to avoid redundant computation due to the misalignment among the connected layers~\cite{1811.09332}. For the second problem,  a straightforward solution consists in directly minimizing the variance of the weights within the group. However, a variance minimization constraint would encourage the value of the weights to be closer rather than the difference in magnitude to be smaller.

In this work, we propose {Variance-Aware Cross-Layer} (VACL) regularization to address these problems. VACL consists of two main components: a cross-layer grouping to enforce channel aligned sparsity across connected layers (see Fig~\ref{fig:motivation}{\color{red}.c}), as well as a novel regularizer to minimize the differences of the magnitudes among parameters within a group. Our experimental results show that the proposed method can effectively prune residual models. For CIFAR-10, it can reduce ResNet110 model size by $79.5\%$ with slightly better accuracy, and ResNeXt-29-8-64 model size by $82\%$ with $<1\%$ accuracy drop. For ImageNet, it achieves a $63.3\%$ pruned ratio with $1.61\%$ top-1 and $0.94\%$ top-5 accuracy drop for ResNet50. As demonstrated in our experiments, the proposed approach significantly outperforms other state-of-the-art methods in terms of overall model size and accuracy.

\section{Related Work}
\textbf{Model Pruning}:
Parameter pruning has a long history in the development of light-weight neural networks. 
One of the earliest successes is the optimal brain surgeon approach \cite{298572}. Optimal brain surgeon analyzes the importance of each parameter in a pre-trained model, keeps only the necessary parameters, and adjusts them to approximate the original accuracy.
Nowadays, sparsity promoting constraints are commonly used to generate sparse models~\cite{Han:2015:LBW:2969239.2969366, Scardapane:2017:GSR:3067301.3067328, 1608.08710, NIPS2016_6504}. These constraints can push many parameters towards zero, so that these parameters no longer have an observable impact on the final accuracy, and can thus be removed. 
Model pruning can be applied in a structural or non-structural manner. Non-structural methods remove individual parameters, which results in sparse convolutional filters that cannot take advantage of fast dense matrix multiplication \cite{Han:2015:LBW:2969239.2969366}. In contrast, structural methods~\cite{NIPS2016_6504, 1708.06519} remove parameters in units of filter or layers. In this case, a pruned model can take advantage of dense matrix computation just as the original model does. $L_1$ norm regularization is commonly used to enforce individual parameter sparsity, and Group Lasso \cite{Yuan06modelselection} is normally used to enforce structural parameter sparsity\cite{Scardapane:2017:GSR:3067301.3067328}.

\textbf{Structural Model Pruning}:
Li \textit{et al.} propose a method to evaluate the importance of each filter and removes filters that do not have a significant impact on accuracy~\cite{1608.08710}. He \cite{He_2017_ICCV} introduced an iterative inference time pruning method. For each layer, it selects and prunes non-representative channels from the model, and then reconstructs the accuracy with remaining channels. Luo \textit{et al.} \cite{Luo_2017_ICCV} measured a filter's importance according to the next layer's statistics and then removed unimportant ones. Gao~\textit{et al.} \cite{1810.05331} proposed a dynamic channel pruning method, which keeps the original model architecture and dynamically skips the computation for unimportant channels. 

\textbf{Structural Sparsity Regularization}:
Based on the Group Lasso regularization proposed by Yuan \cite{Yuan06modelselection}, various sparsity regularizations have been investigated to improve the structured sparsity. In the work of Structured Sparsity Learning (SSL) \cite{NIPS2016_6504}, weight groups in different scales were defined to enforce structured sparsity. It learns a compact model architecture by adjusting filter shapes, filter numbers, or model depth. Alvarez and Salzmann~\cite{NIPS2016_6372} used group sparsity regularization to automatically learn the number of neurons during training to obtained a compact model. Similarly, Scardapane \cite{Scardapane:2017:GSR:3067301.3067328} introduced Sparse Group Lasso, which is a combination of $L_1$ and Group Lasso. It improves structural weight sparsity at the group level, as well as individual weight sparsity within a group. Wen~\textit{et al.} \cite{NIPS2016_6504} developed a structured sparsity learning method to adjust filter shapes, layer channel numbers, and model depth to obtain the pruned model structure. MorphNet \cite{Gordon_2018_CVPR} uses Group Lasso to preserve network topology while learning the model structure. Lebedev \cite{Lebedev_2016_CVPR} made use of group-sparsity regularization and group-wise brain damage to speedup convolutional layers. In this paper, we propose VACL as an extension of Group Lasso. Our method first enforces aligned sparsity across multiple connected layers, and then encourage parameter similarity to improve channel pruning efficiency.


\begin{figure*}[!t]
\begin{center}
\includegraphics[width=0.7 \linewidth]{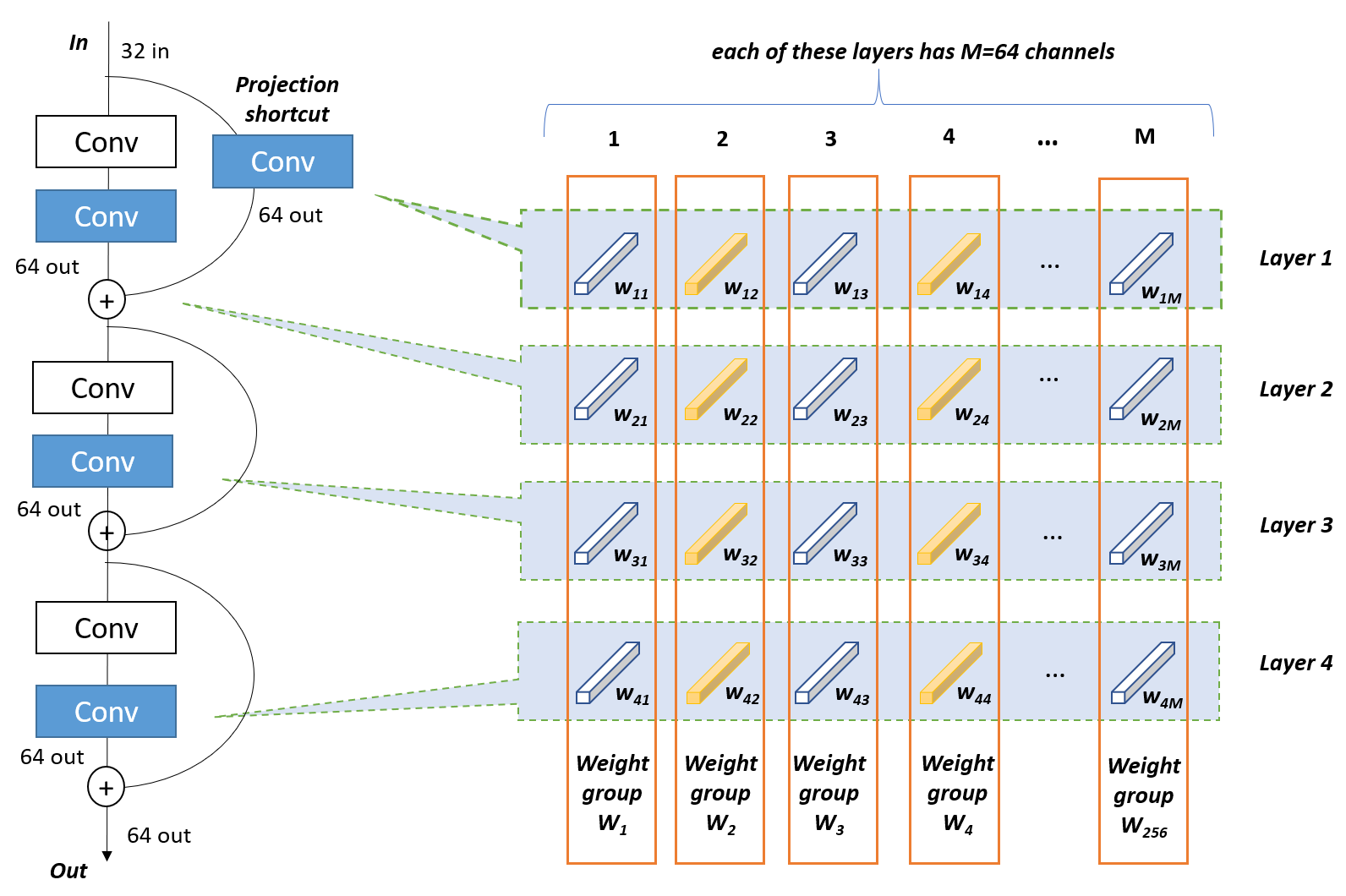}
\end{center}
\vspace{-0.25cm}
   \caption{Proposed Cross-Layer grouping for element-wise connected layers. }
\label{fig:def_vacl}
\vspace{-0.5cm}
\end{figure*}

\section{VACL Regularization}
\label{sect:method}
In this section we introduce our novel regularizer to improve the structural sparsity of residual layers. To this end, we first introduce the grouping approach namely Cross-Layer grouping, and then the proposed Variance-Aware penalty.

\subsection{Cross-Layer Grouping}
Given $N$ training input-output pairs  $(x_j, y_j)$, we can formulate the cost function to train a deep neural network as:
\begin{equation}
\mathcal L(X, \textbf{W}) = \sum_{j=1}^N E (y_j, f(x_j, \textbf{W})) + \lambda R(\textbf{W}),
\label{eq:def_loss}
\end{equation}
\noindent where $f$ represents a generic deep neural network and $\textbf{W}$ all the parameters of such a network, $E(\cdot , \cdot )$ is a prediction (supervised) loss, such as cross-entropy for classification, $R(\cdot)$ is a regularization penalty term acting on the network parameters and $\lambda$ is the strength of the regularizer. Examples or penalty terms are the $L_2$ norm encouraging small magnitude weights or the $L_1$ norm (lasso) which encourages parameter sparsity.

\begin{figure*}[t!]
\begin{center}
\begin{tabular}{ccccc}
     \hspace{-0.5cm}\includegraphics[width = 0.22\textwidth]{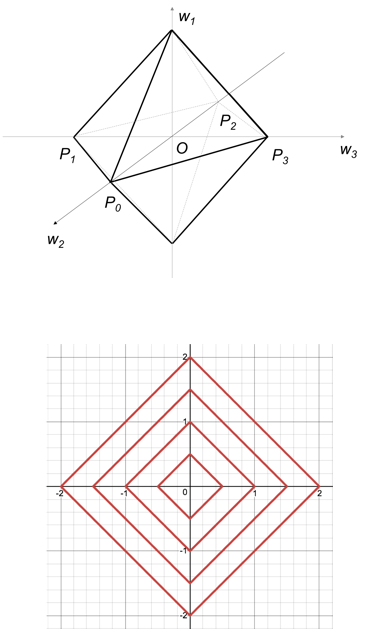} &
     \hspace{-0.5cm}\includegraphics[width = 0.22\textwidth]{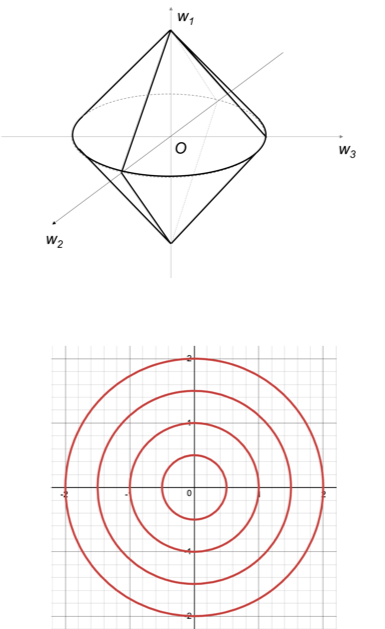} &
     \hspace{-0.5cm}\includegraphics[width = 0.22\textwidth]{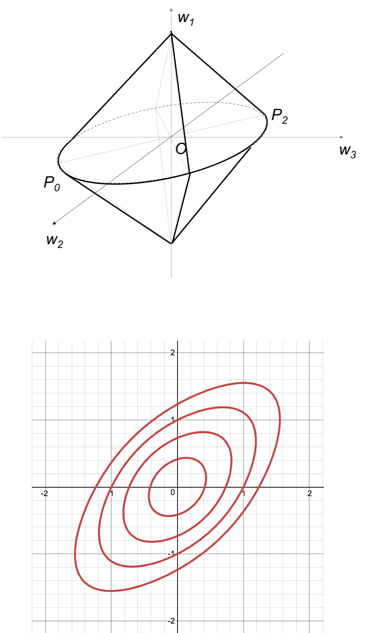} &
      \hspace{-0.5cm}\includegraphics[width = 0.22\textwidth]{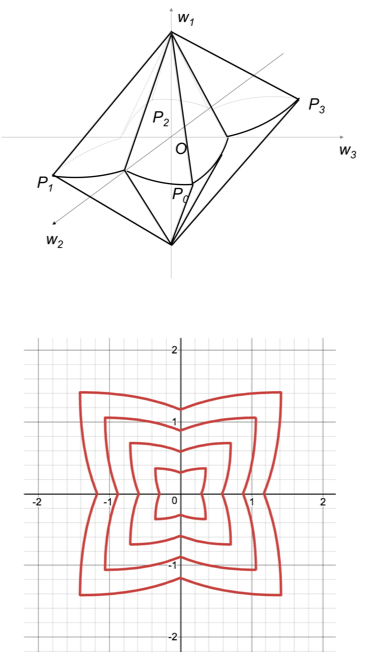}\\
      \hspace{-0.5cm}(a)&\hspace{-0.5cm}(b)&\hspace{-0.5cm}(c)&\hspace{-0.5cm}(d)\\ 
\end{tabular}
\end{center}
\vspace{-0.15cm}
   \caption{Feasible sets for different penalties for the case of three (groups of) variables. a) $L_1$ norm; b) Group-Lasso; c) Group-Lasso with Variance; and d) Our proposal to minimize the differences of the magnitudes among parameters within a group as defined in Eq.~\ref{eq:def_2-norm} (\textbf{Ours}).} 
\label{fig:contours}
\vspace{-0.5cm}
\end{figure*}

Group Lasso is an extension of the lasso penalty commonly used to encourage groups of parameters to become zero (non-zero) simultaneously~\cite{NIPS2016_6372}. In the particular case of a deep neural network, a common approach is to consider each neuron as a group of parameters. Given $\textbf{W}$ the set of weights divided into $Q$ groups (e.g., number of neurons in the network), the Group Lasso penalty is defined as:
\begin{equation}
R_{\small{gl}}(\textbf{W}) = \sum_{i=1}^{Q}\sqrt{p_i}\cdot||W_i||_2,
\label{eq:def_group_lasso_2}
\end{equation}
\noindent where $p_i$ is the size of the $i$-th group. The constrained region for the Group Lasso over two groups of weights $W_1=\{w_1\}$ and $W_2=\{w_2, w_3\}$ is shown in Fig.~\ref{fig:contours}{\color{red}.b}. As can be seen, for weights within the same group ($w_2$ and $w_3$), the $L_2$ norm penalty treats every direction equally and does not induce sparsity. In contrast, the $L_1$ norm penalty enforces sparsity between groups of weights ($W_1$ and $W_2$).

Group Lasso has been widely applied to prune neural networks taking advantage of the network structure and producing models suitable for dense matrix multiplications~\cite{NIPS2016_6504, NIPS2016_6372}. However, Group Lasso efficiency to prune neurons drops in networks using residual connections (i.e., skip-connections and element-wise additions). In those cases, Group Lasso produces sparsity patterns in the connected layers that are not aligned and therefore, corresponding neurons cannot be effectively removed after the addition operation (see Fig.~\ref{fig:motivation}). To solve that problem, we propose Cross-Layer grouping as an extension of Group Lasso. Our proposal considers all the layers interacting (connected) in skip-connections and aggregates in a single group whose parameters from neurons indexed by the same index, as shown in Fig. \ref{fig:def_vacl}. In Cross-Layer grouping, a group of weights is defined as:
\begin{equation}
W_i = \cup \textbf{w}_{li}
\label{eq:def_group_weight_union}
\end{equation}
\noindent where $l\in[1,L]$ represents the $l$-th layer in a set of layers that are element-wise connected, $i\in[1,M]$ is the number of neurons in each of these layers and $\textbf{w}_{li}$ the set of parameters of the $i$-th neuron in the $l$-th layer. As we will demonstrate in our experiments, using this novel definition of groups, structural sparsity constraints such as Group Lasso can be effectively applied to enforce group level sparsity in residual networks, so that the $i^{th}$ filters of all layers in $\textit{L}$ could be either significant, or redundant.

\subsection{Variance-Aware Regularization}
In the previous section we have defined our approach to group weights within connected layers to subsequently apply sparsity regularizers such as Group Lasso. In the Group Lasso formulation, groups are equally penalized due to the weighting accordingly to their size $\sqrt{p_i}$. Within each group, weights (parameters) are pushed towards zero with the same strength independently of their magnitude. Therefore, it is possible that most weights within a group are very small, but some are very large, which results in the entire group not being removed, and thus inefficient sparsity results. This is particularly relevant when the size of the groups increases and when the parameters within a group come from different layers, hence different learning pace. To address this problem, we propose variance-regularization, a penalty term that constraints the variance among the weights in each group.

\JM{Detail the groups and the variables. Be consistent with naming. Remove Var and refer as r.}
A straightforward method to reduce the variance of the weights within a group is to directly minimize the variance of the weights:
\begin{equation}
\begin{split}
\text{Var}(W_i)=||W_i-\overline{W_i}\cdot \mathbbm{1}||_2^2.
\end{split}
\label{eq:Var}
\end{equation}
The constrained region for this penalty is shown in Fig.~\ref{fig:contours}{\color{red}.c}. As shown, this term encourages the value of the weights to be closer rather than their magnitude ( $|w_2|$ be close to $|w_3|$ in the example). Therefore, we propose a variant aiming to reduce the variance of the absolute values of the weights in the group. More precisely, our Variance-Aware penalty is defined as:
\begin{equation}
\begin{split}
r_{\small{var}}(W_i) =|||W_i|-\overline{|W_i|}\cdot \mathbbm{1}||_2.
\end{split}
\label{eq:def_2-norm}
\end{equation}

The contours and constrained region for this penalty are shown in  Fig.~\ref{fig:contours}{\color{red}.d}. As shown, as our approach constraints the absolute value of the weights, each contour has four sharp corners in the two diagonal directions. In each of these directions, the penalty has minimum weight magnitude variance.


\subsection{Variance-Aware Cross-Layer Regularization}
In this section, we combine the Variance-Aware penalty with the Cross-Layer grouping described above. This let us define our Variance-Aware Cross-Layer (VACL) regularization as:
\begin{equation}
\begin{split}
R_{\small{vacl}}(\textbf{W}_g) = \sum_{g \in \bf G} \sum_{i=1}^M\sqrt{p_i^g}\bigg{[} ||W_i^g||_2 +  |||W_i^g|-\overline{|W_i^g|}\cdot \mathbbm{1}||_2 \bigg{]},
\end{split}
\label{eq:def_vacl}
\end{equation}
\noindent where $\textbf{W}_g$ is the set of weights that can be grouped using Cross-Layer grouping; $W_i^g$ refers to the weights of the $i^{th}$ filter in $g-th$ group, and $p_i^g$ is the number of weights in $W_i^g$. This novel regularization term encourages the magnitude of weights within a group to be close to each other. 


In practice, a deep neural network consists of layers that are element-wise connected and other layers that are not. We define the regularization over all the weights as the combination of a Variance-Aware Cross-Layer term for the first type and a group sparsity term for the rest of layers. That is, the cost function to train a network becomes, 
\begin{equation}
\mathcal L(X, \textbf{W}) = \sum_{j=1}^N E (y_j, f(x_j, \textbf{W})) + \lambda(R_{gl}(\textbf{W}_s) + R_{vacl}(\textbf{W}_g)),
\label{eq:def_total_loss}
\end{equation}
\noindent where $\textbf{W}_g$ is again those weights corresponding to layers that are connected through skip connections and, $\textbf{W}_s$ refers the weights from the rest of layers, and $\textbf{W}=\textbf{W}_g\cup\textbf{W}_s$.

\subsection{Model Pruning Algorithm} We adopt the pruning pipeline proposed by Han~\cite{Han:2015:LBW:2969239.2969366}. We first train a sparse model using a sparsity promoting regularization such as the proposed VACL, and then prune the model at the filter level. Subsequently, we fine-tune the model with an optional $L_2$ regularization. If the resulting model (after fine-tuning) does not meet the expected trade-off between accuracy and model size, we repeat the first two steps. We call the first two steps one \textbf{train-prune stage}.  

To prune the model, we make use of a filter importance criterion defined as:
 \begin{equation}
 \mathcal{I}_{li} = \frac{ |\textbf{w}_{li}| }{ \sum_i{|\textbf{w}_{li} }| },
 \end{equation}
\noindent where $|\textbf{w}_{li}|$ is the $l_1$ norm of the $i$-th filter in the $l$-th layer. Given this criterion and a threshold $\tau$, we decide whether a filter is kept or removed,
\begin{equation}
 \textbf{w}_{li} =
  \begin{cases} 
    \textbf{w}_{li}, & \text{if $\mathcal{I}_{li} \geq \tau$.} \\
    \text{removed }, & \text{otherwise. }
  \end{cases} 
\end{equation}

\section{Experiments}
In this section, we demonstrate the benefits of our Variance-Aware Cross-Layer (VACL) regularization on image classification on the popular datasets CIFAR10, CIFAR100 \cite{CIFAR-10} and ImageNet ILSVRC-2012 \cite{imagenet}. CIFAR10 and CIFAR100 contain 50,000 training images and 10,000 test images of 10 and 100 different classes, respectively. The images are of size $32\times32$. ILSVRC-2012 subset of ImageNet consists of 1000 categories, with 1.2 million training images and 50,000 validation images. In this case, we use the standard pre-processing where images are resized to a resolution of $224\times224$. All our experiments are conducted on a NVIDIA DGX-1 using Tensorflow\cite{tensorflow2015-whitepaper} and Keras \cite{chollet2015keras}. 

In the following sections, we first provide a set of ablation studies to more thoroughly analyze the influence of the proposed regularization technique compared to its counterparts, then, we provide comparison to state-of-the-art methods. In a third experiment, we show experimental results on using the pruned models to train networks from scratch and finally, we analyze the ability to transfer across domains of the resulting models. As baseline models we consider the models reported in~\cite{He_2016_CVPR} and~\cite{Xie_2017_CVPR}. When comparing to state-of-the-art pruning methods, we choose those works whose reported results have: i) a pruned ratio (percentage of weights removed) higher than $10\%$; and ii) a top-1 accuracy drop lower than $10\%$. 

\begin{figure}[t!]
\begin{center}
\includegraphics[width=0.85 \columnwidth]{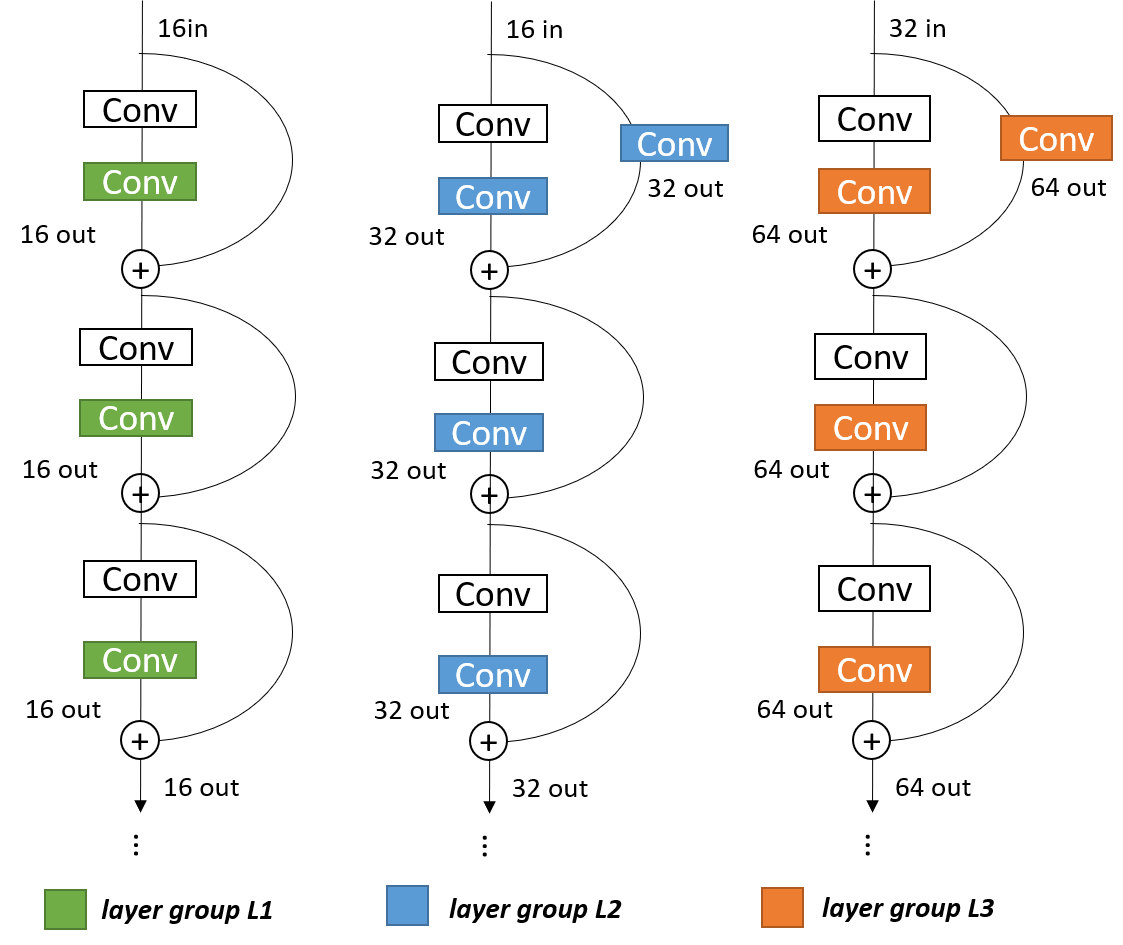}
\end{center}
   \caption{Grouping of element-wise connected layers for a ResNet-110 on CIFAR10.}
\label{fig:cl_resnet110}
\end{figure}

In our experiments, we obtain regularization weight $\lambda$ by grid search at the initial training stage, and we set pruning threshold $\tau$ as 0.0001. 

\begin{figure}[t!]
\begin{center}
\includegraphics[width=0.85\linewidth]{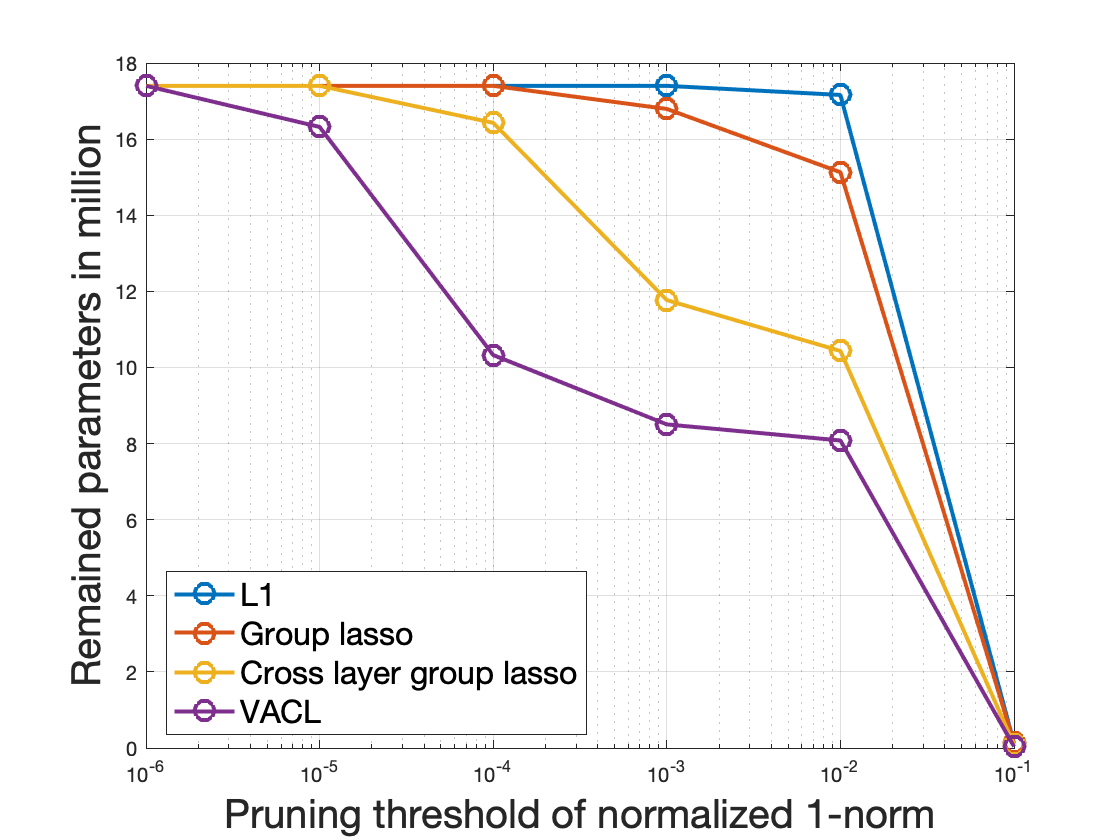}
\end{center}
   \caption{\textbf{Parameter sensitivity for ResNet-110 on CIFAR-10}. Model size as a function of the pruning threshold $\tau$ for different sparsity promoting penalties. All these models have a similar accuracy in the initial training stage of $\simeq 0.91\%$. }
\label{fig:pruning_threshold}
\vspace{-0.25cm}
\end{figure}

\subsection{Parameter sensitivity and ablation studies}
\textbf{Comparison to other sparsity regularizers.} The goal of the first experiment is to compare the behavior of different sparsity regularization strategies. Specifically, we compare $L_1$, Group Lasso, Cross-Layer Group Lasso, and VACL for training a ResNet-110 architecture on CIFAR10. A ResNet-110 model comprises three element-wise connected groups of layers as shown in Fig.~\ref{fig:cl_resnet110}. We use this grouping for Cross-Layer Group Lasso and VACL. The value of $\lambda$ for each strategy is set using grid search so as to achieve a top-1 accuracy of around $91\%$. We also analyze the sensitivity of each regularizer to the pruning threshold, $\tau$. Fig.~\ref{fig:pruning_threshold} summarizes this analysis.

In Fig.~\ref{fig:pruning_threshold}, we can observe that using the proposed VACL leads to more compact models, especially when compared to lasso and Group Lasso. As shown, the lasso penalty is not able to effectively reduce many filters in these layers as it only operates at the parameter level. In contrast, our proposal method quickly starts reducing the number of parameters as the threshold increases. As expected, for a large value of $\tau$, all filters are removed and therefore the model is completely pruned. 


\begin{figure}[t!]
\begin{center}
\begin{tabular}{cc}
\vspace{-0.5cm}
 \centering\hspace{-0.15cm}\begin{sideways}\centering \hspace{1.1cm}\scriptsize{$L_1$}\end{sideways}& \hspace{-0.5cm} \includegraphics[width =0.98\columnwidth]{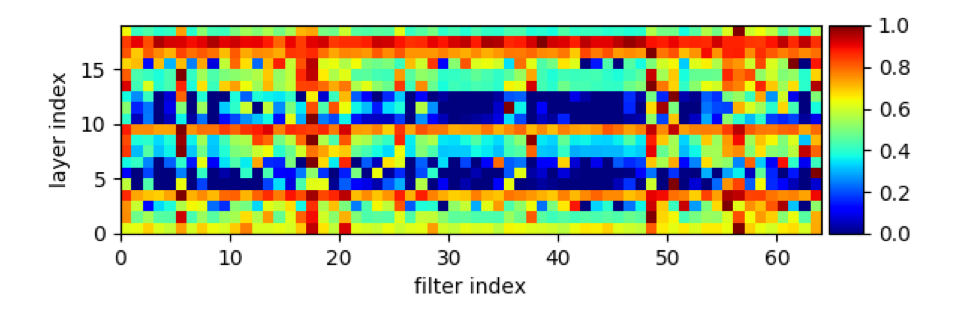}\\ \vspace{-0.5cm}
 \centering\hspace{-0.15cm}\begin{sideways}\centering \hspace{0.85cm}\scriptsize{Group Lasso}\end{sideways}& \hspace{-0.5cm}\includegraphics[width = 0.98\columnwidth]{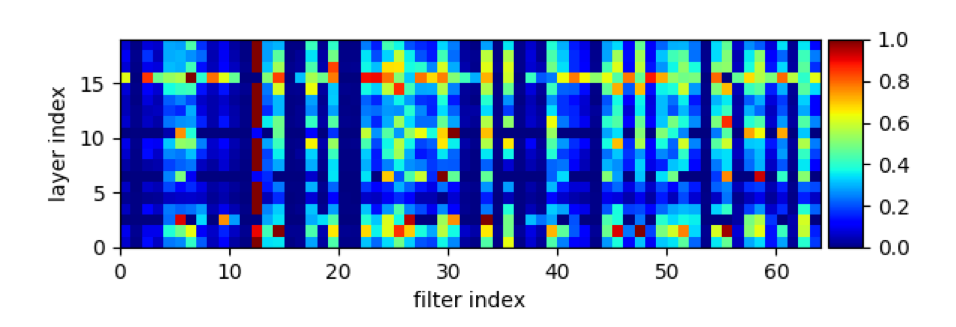}\\ \vspace{-0.5cm}
 \centering\hspace{-0.15cm}\begin{sideways}\centering \hspace{1.1cm} \scriptsize{CL-GL}\end{sideways} &\hspace{-0.5cm}\includegraphics[width = 0.98\columnwidth]{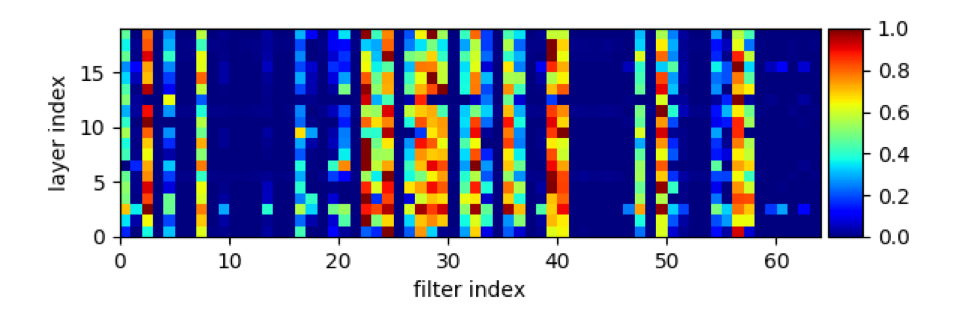}\\ \vspace{-0.5cm}
 \centering\hspace{-0.15cm}\begin{sideways}\centering \hspace{1.1cm} \scriptsize{VACL}\end{sideways}
 &\hspace{-0.5cm}\includegraphics[width = 0.98\columnwidth ]{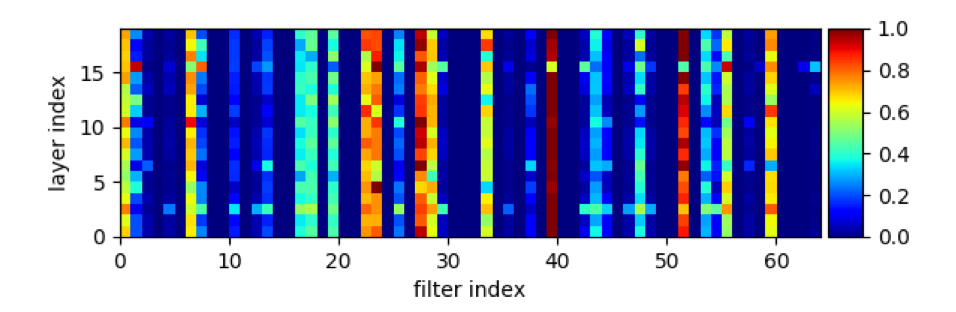}\\ \vspace{-0.05cm}
\end{tabular}
\end{center}
\vspace{-0.25cm}
   \caption{Heatmaps of filter importance for $L_1$, Group Lasso, Cross-Layer Group Lasso (CLGL) and the proposed Variance-Aware Cross-Layer (VACL) penalties. Importance of a filter is measured by the normalized $L_1$ norm of the filter (see text for details). In each plot, $x$-axis represents the filter index in a layer, and the $y$-axis represents the layer index within the group of element-wise connected layers. \JM{Ideally vertical to better visualization of the alignment. Improve caption, details and a quick take-home message -- Also numbers}  }
\label{fig:heatmap}
\vspace{-0.3cm}
\end{figure}

Fig. \ref{fig:heatmap} shows a comparison of the sparsity results for the third group of element-wise connected layers (layer group L3 in Fig.~\ref{fig:cl_resnet110}). This grouping comprises 19 element-wise connected layers (rows), each with 64 filters (columns). The importance of each filter is represented using the normalized $||\cdot||_1$. As shown, Group Lasso (Fig.~\ref{fig:heatmap}{\color{red}.b}) achieves better structural sparsity than $L_1$ (Fig.~\ref{fig:heatmap}{\color{red}.a}). That is, using Group Lasso, we obtain a larger number of unimportant columns (the same filter in all the layers is unimportant). Results are even better when using Cross-layer Group-Lasso (see Fig.~\ref{fig:heatmap}{\color{red}.c}) where we can observe a better alignment in the vertical direction. Furthermore, comparing Cross-Layer Group Lasso to VACL (Fig.~\ref{fig:heatmap}{\color{red}.c} and Fig.~\ref{fig:heatmap}{\color{red}.d}) we can observe that filters within a layer have lower variance which leads to a better pruning efficiency and, more importantly, reduces the sensitivity of the algorithm to the pruning threshold as shown in Fig.~\ref{fig:pruning_threshold}. From these results, we can conclude that using the proposed VACL regularization term leads to more compact models when compared to other sparsity promoting regularizers.

\begin{figure}[t!]
\begin{center}
\begin{tabular}{cc}
\hspace{-0.5cm}\includegraphics[width = 0.55\linewidth]{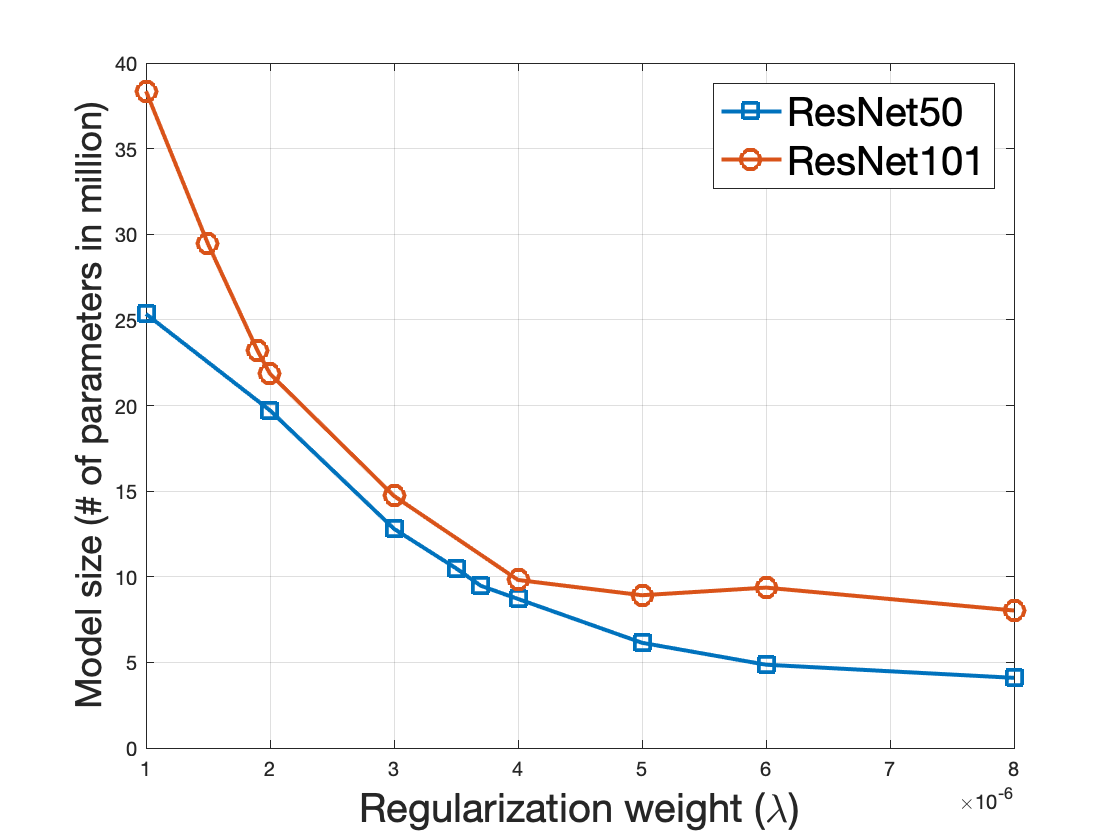}&
\hspace{-0.7cm}\includegraphics[width = 0.55\linewidth]{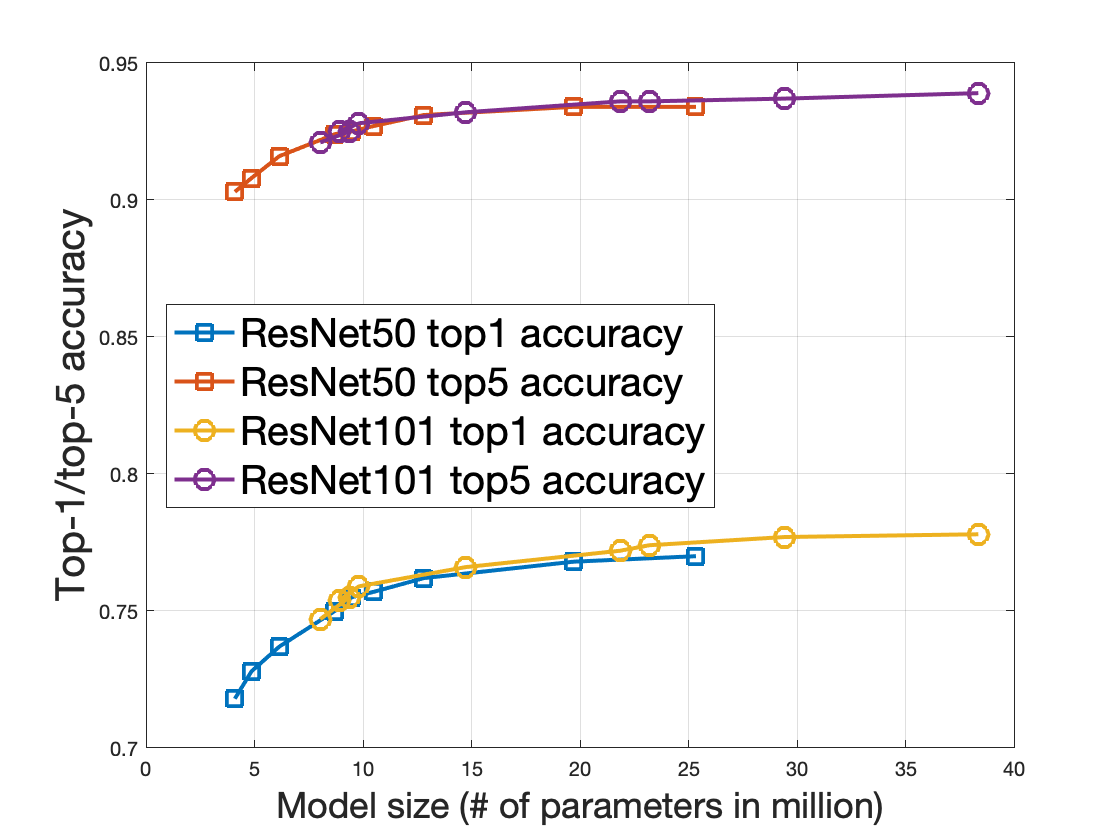} \\ \hspace{-0.5cm}(a)& \hspace{-0.7cm}(b)
\end{tabular}
\end{center}
\vspace{-0.25cm}
   \caption{\textbf{Parameter sensitivity for Resnet-50/101 on ImageNet.} (a) Model size as a function of $\lambda$, the regularization strength. (b) Accuracy (top-1 and top-5) as a function of the model size.}
\label{fig:variant_reg_weight}
\vspace{-0.25cm}
\end{figure}

\textbf{Sensitivity to regularization strength.} We now focus on analyzing the effect of the proposed regularizer on the accuracy and the model size. To this end, we make use of ResNet-50 and ResNet-101 on ImageNet, and measure the impact of varying $\lambda\in[1e-6, 8e-6]$ in Eq.~\ref{eq:def_total_loss}. Each data point is obtained by training a model, pruning it, and fine-tuning it with a $L_2$ regularization with $\lambda=1e^{-4}$. Fig.~\ref{fig:variant_reg_weight} shows the summary of results for this analysis. As $\lambda$ increases, the pruned ratio increases significantly while the top-1 accuracy drops at a slower rate. Beyond a certain value of $\lambda$, the top-1/top-5 accuracy starts dropping at a much higher rate. \JM{Add a short conclusion and a benefit of VACL}

\textbf{Stability comparison over multiple trials.} In the last ablation analysis, our goal is to analyze the stability of the results for models trained using the proposed VACL compared to other sparsity promoting penalties. For this experiment, we make use of a ResNet-110 model and a ResNet-101 model on CIFAR10 and ImageNet respectively. Specifically, we repeat the same experiment $5$ times using the same parameters as for previous experiments and report the variation in model size over the trials. As in previous experiments, all models are trained to achieve a similar top-1 accuracy $91\%$. Table \ref{tab:stability_analysis} shows a summary of results for this experiment. 

As shown in Table \ref{tab:stability_analysis}, using a Cross-Layer weight grouping penalty reduces significantly the average model size ($\mu$) when compared to group-lasso. In addition, the reduction in both average model size and deviations is even better if the proposed VACL penalty is used. From these results, we can conclude that the proposed VACL penalty not only produces more compact models compared to other penalties, but also improves the stability of the results in terms of model size.

\begin{table}[t!]
\footnotesize
\begin{center}
\begin{tabular}{c|c||c|} 
\cline{2-3}
 & \multicolumn{2}{c|}{$\#$ params (in M)}\\ \cline{2-3}
 & CIFAR-10 &ImageNet\\  \hline
GL & $0.53 \pm 0.17$ &$23.27 \pm 0.52$ \\
CLGL & $0.68 \pm 0.03$ &$22.10\pm 0.64$\\
VACL-GL (ours) & $0.60\pm 0.01$ &$22.08\pm 0.12$\\\hline
\end{tabular}
\end{center}
\caption{The proposed VACL regularization improves the stability of weight sparsity over multiple runs. mean $\pm$ std. dev. of 5 runs }
\label{tab:stability_analysis}
\vspace{-0.25cm}
\end{table}

\begin{table*}[t!]
\footnotesize
\begin{center}
\begin{tabular}{|c|c|c|c|c|c|c|} 
\hline
L\# & Method & Baseline Error (\%) & Pruned Error (\%) &  Error Inc (\%) & Params \# & Pruned(\%)  \\
\hline\hline
56 & Soft Filter \cite{1808.06866} & 6.97 & 6.65$\pm$0.31 & -0.32 & 0.51M & 40\% \\
56 & \textbf{VACL-GL ($\lambda=1e^{-5}$)} & 6.97 & \textbf{6.66} & \textbf{-0.31} & \textbf{0.30M} & \textbf{65.7\%} \\
56 & Li-ScratchB\cite{1802.00124} & 6.97 & 6.91$\pm$0.14 & -0.06 & 0.73M & 13.7\% \\
56 & Li\cite{1608.08710} & 6.97 & 6.90 & -0.07 & 0.73M & 13.7\%  \\
\hline\hline
110 & Soft Filter \cite{1808.06866} & 6.43 & 6.14$\pm$0.21 & -0.29 & 1.19M & 30\% \\
110 & \textbf{VACL-GL ($\lambda=8e^{-6}$)} & 6.43 & 6.35 & -0.08 &  \textbf{0.36M} & \textbf{79.5\%} \\
110 & Li-ScratchB\cite{1802.00124} & 6.43 & 6.40$\pm$0.25 & -0.03 & 1.16M & 32.4\% \\
110 & Li\cite{1608.08710} & 6.43 & 6.70 & 0.27 & 1.16M & 32.4\%  \\
\hline
\end{tabular}
\end{center}
\caption{\textbf{ResNet on CIFAR-10, comparison to existing approaches.} [Error Increase \%] denotes the absolute error increase, and a negative value indicates an improved model accuracy; [Pruned \%] denotes the reduction of model parameters.}
\label{tab:cifar10}
\vspace{-0.25cm}
\end{table*}


\textbf{Sensitivity to train-prune iterations.} We also investigate the effect of applying VACL regularizaton over multiple train-prune iterations. To this end, we make use of a ResNet-110 model on CIFAR10. We train the model over multiple train-prune stages varying the type of regularization during training ($L_1$ and VACL) with a fine-tune process with no regularization after each stage. As a baseline, we consider the model trained for 5 stages using $L_1$ norm as regularization. We compared the performance of that baseline model to the performance of applying VACL during the third and after the fifth stage. Table~\ref{tab:6l1_vacl} shows the summary of these results. As shown, after the $2^{nd}$ stage, the accuracy of the baseline starts to fluctuate with an insignificant improvement on the pruning ratio over consecutive iterations. In contrast, if we apply VACL to the element-wise connected layers after the $2^{nd}$ iteration, the model size is reduced from 0.54M to 0.42M with a 0.34\% improvement in accuracy. This improvement is even larger if we apply VACL at the $5^{th}$ iteration. In this case, the model size can be reduced from 0.48M to 0.4M with a 0.32\% improvement in accuracy. From these results, we can conclude that our proposed VACL regularization method is effective to improving sparsity to models that have been already trained and pruned leading to better accuracy and a significant reduction in model size.


\begin{table}[t!]
\footnotesize
\begin{center}
\begin{tabular}{|c|c|c|c|c|} 
\hline
Train-prune &  \multicolumn{2}{c|}{Acc. (\%) / $\#$ Params (M) } \\ \cline{2-3}
iteration  &  $L_1$ & VACL \\\hline
1 & 0.9251 / 0.607 &  -- \\
2 & 0.9298 / 0.542 &  -- \\
3 & 0.9268 / 0.519 &  0.9332 / 0.424 \\
4 & 0.9307 / 0.488 &  -- \\
5 & 0.9309 / 0.482 & 0.9341 / 0.401 \\
\hline
\end{tabular}
\end{center}
\vspace{-0.25cm}
\caption{\textbf{ResNet-110 on CIFAR10, sensitivity to train-prune iterations.} We iterate over train-prune stages and analyze the performance of the models after each step. Applying our VACL approach at any stage leads to more compact models with slight accuracy improvement.}
\label{tab:6l1_vacl}
\vspace{-0.25cm}
\end{table}

\subsection{Comparison to existing approaches}

\begin{table*}[t!]
\footnotesize
\begin{center}
\begin{tabular}{|c|c|c|c|c|c|c|c|} 
\hline
Dataset & Network & Method & Baseline Error (\%) & Pruned Error (\%) &  Error Inc (\%) & Params \# & Pruned\%  \\
\hline\hline
CIFAR10 & 29-8-64 & \textbf{VACL-GL ($\lambda=3e^{-6}$)} & 3.65 & 4.61 & 0.96 & \textbf{6.1M} & \textbf{82\%} \\
CIFAR10 & 29-8-64 & Zhang's method\cite{Zhang_2018_CVPR_Workshops} & 3.65 & 4.09 & 0.44 & 9.1M & 73\% \\
\hline\hline
CIFAR100 & 29-8-64 & \textbf{VACL-GL ($\lambda=2e^{-6}$)} & 17.77 & 19.76 & 1.99 & 12.1M & 65\% \\
\hline
\end{tabular}
\end{center}
\caption{\textbf{ResNeXt on CIFAR-10 and CIFAR-100, comparison to existing approaches.} [Error Increase \%] denotes the absolute error increase, and a negative value indicates an improved model accuracy; [Pruned \%] denotes the reduction of model parameters.}
\label{tab:cifar100}
\end{table*}


We now compare our results to those provided by existing methods on CIFAR and ImageNet.
\textbf{\\CIFAR.} We evaluate our VACL-GL regularization on ResNet-50, ResNet-110 and ResNetXt on CIFAR datasets. For ResNet, all models are trained for 200 epochs with the initial learning rate is 0.001, divided by 10 at 80, 120, 150 epochs, and divided by 2 at 180 epochs. Other hyper-parameters are the same as in \cite{He_2016_CVPR}. ResNet models are trained using VACL-GL for 7 train-prune stages and then, fine-tuned with no regularization. ResNeXt models are trained for 300 epochs, with the initial learning rate set to 0.1, and divided by 10 at 100, 180 and 250 epochs. The rest of parameters are the same as in~\cite{Xie_2017_CVPR}. In this case, the model is trained using VACL-GL for a single train-prune stage and then, fine-tuned with $L_2$ regularization ($\lambda=1e^{-4}$). 

We compare our results to those reported by Li's method~\cite{1608.08710}\cite{1802.00124} and by Soft Filter~\cite{1808.06866} for ResNet in Table~\ref{tab:cifar10} and those reported by Zhang's method~\cite{He_2017_ICCV} for ResNetXt in Table~\ref{tab:cifar100}. As shown, ResNet models trained using the proposed method achieve up to $0.31\%$ accuracy improvement with a pruning ratio up to $65.7\%$-$79.5\%$ when compared to the baseline. Moreover, our results significantly outperform existing methods in terms of the pruned ratio for a better or comparable accuracy. For ResNeXt models, the proposed method outperforms the pruning ratio of Zhang's method by achieving $>80\%$ pruned ratio, with less than $1.0\%$ accuracy drop.
\textbf{\\ImageNet.} Results on the ImageNet dataset are shown in Table \ref{tab:imagenet}. ResNet50/110 are trained for 150 epochs with initial learning rate = 0.128, divided by 10 at 45, 90, 125 epochs. Other hyper-parameters are the same as in \cite{He_2016_CVPR}.  
Each experiment follows the same train-prune-finetune workflow: train a sparse model, prune it, and fine-tune it with $L_2$ regularization with $\lambda_{l2}=1e^{-4}$.

As the results show, compared to other state-of-the-art method, the proposed method achieves significantly higher pruned ratio, with better or comparable accuracy. 
For ResNet50, VACL-GL achieves $63.3\%$ pruned ratio with $<2\%$ top-1 accuracy drop; and for ResNet110, VACL-GL achieves $48.1\%$ pruned ratio with $<1\%$ top-1 accuracy drop compared with baseline models.

\begin{table*}[t!]
\footnotesize
\begin{center}
\begin{tabular}{|c|p{3.5cm}|c|c|c|c|c|c|c|c|} 
\hline
$L\# $
& Method 
& \multicolumn{1}{p{1.2cm}|}{\centering Top-1 \\ error \\ baseline (\%)}
& \multicolumn{1}{p{1.2cm}|}{\centering Top-1 \\ error \\ pruned (\%)}
& \multicolumn{1}{p{1.2cm}|}{\centering Top-1 \\ error \\ increase (\%)}
& \multicolumn{1}{p{1.2cm}|}{\centering Top-5 \\ error \\ baseline (\%)}
& \multicolumn{1}{p{1.2cm}|}{\centering Top-5 \\ error \\ pruned (\%)}
& \multicolumn{1}{p{1.2cm}|}{\centering Top-5 \\ error \\ increase (\%)}
& Param\#
& Pruned\%  \\
\hline\hline
50 & \textbf{VACL-GL} ($\lambda=3.7e^{-6}$) & 22.85 & \textbf{24.53} & \textbf{1.68} & 6.71 & \textbf{7.27} & \textbf{0.56} & \textbf{9.4M} & \textbf{63.3\%} \\
50 & Soft Filter\cite{1808.06866} &  22.85 & 25.39 & 2.54 & 6.71 & 7.94 & 1.23 & 17.9M & 30\% \\
50 & \scriptsize{Sparse Structure Selection-32\cite{Huang_2018_ECCV}} &  22.85 & 25.82 & 2.97 & 6.71 & 8.09 & 1.38 & 18.6M & -- \\
50 & NISP\cite{Yu_2018_CVPR} &  22.85 & 27.33 & 4.48 & --  & -- & -- & 11.2M & 56.2\%  \\
50 & ThiNet70\cite{Luo_2017_ICCV} &  22.85 & 27.96 & 5.11 & 6.71 & 9.33 & 2.62 & 16.9M & 30\%  \\
50 & \scriptsize{Sparse Structure Selection-26\cite{Huang_2018_ECCV}} &  22.85 & 28.18 & 5.33 & 6.71 & 9.21 & 2.62 &  15.6M & -- \\
50 & ThiNet50\cite{Luo_2017_ICCV} &  22.85 & 28.99 & 6.14 & 6.71 & 9.98 & 3.27 & 12.4M & 50\%  \\
\hline\hline
101 & Soft Filter\cite{1808.06866} & 21.75 & \textbf{22.49} & \textbf{0.74} & 6.05 & \textbf{6.29} & \textbf{0.14} & 31.3M & 30\% \\
101 & \textbf{VACL-GL} ($\lambda=2e^{-6}$)  & 21.75 & 22.81 & 1.06 & 6.05 & 6.39 & 0.34 & \textbf{21.9M} & \textbf{50.9\%} \\
\hline
\end{tabular}
\end{center}
\caption{\textbf{ResNet on ImageNet, comparison to existing approaches.} [Error Increase \%] denotes the absolute error increase, and a negative value indicates an improved model accuracy; [Pruned \%] denotes the reduction of model parameters.\JM{Seems like captions are redundant.. are they necessary?}}
\label{tab:imagenet}
\end{table*}
\begin{table*}[t!]
\footnotesize
\begin{center}
\begin{tabular}{|c|c|c|c|c|c|c|c|c|c|} 
\hline
$L\# $
& Method 
& \multicolumn{1}{p{1.2cm}|}{\centering Top-1 \\ error \\ baseline (\%)}
& \multicolumn{1}{p{1.2cm}|}{\centering Top-1 \\ error \\ pruned (\%)}
& \multicolumn{1}{p{1.2cm}|}{\centering Top-1 \\ error \\ increase (\%)}
& \multicolumn{1}{p{1.2cm}|}{\centering Top-5 \\ error \\ baseline (\%)}
& \multicolumn{1}{p{1.2cm}|}{\centering Top-5 \\ error \\ pruned (\%)}
& \multicolumn{1}{p{1.2cm}|}{\centering Top-5 \\ error \\ increase (\%)}
& Param\#
& Pruned\%  \\
\hline\hline
50 & VACL-GL ($\lambda=3.7e^{-6}$)&  22.85 & 24.46 & 1.61 & 6.71 & 7.65 & 0.94 & 9.4M & 63.3\% \\
50 & ThiNet70 \cite{1802.00124} & 22.85 & 24.88 & 2.03 & -- & -- & -- & 17.9M & 30\% \\
50 & ThiNet50 \cite{1802.00124} & 22.85 & 26.10 & 3.25 & -- & -- & -- & 12.8M & 50\% \\
\hline
101 & VACL-GL ($\lambda=2e^{-6}$)& 21.75 & 22.65 & 0.90 & 6.05 & 6.36 & 0.31 & 21.9M & 50.9\%\\
\hline
\end{tabular}
\end{center}
\vspace{-0.15cm}
\caption{\textbf{ResNet on ImageNet, comparison to training a pruned model form random initialization}. Baselines models are taken from Table~\ref{tab:imagenet}. Our results indicate that training the pruned model from scratch achieves a competitive accuracy when compared to the original train-prune-fine-tune process. Compared to existing approaches, our method leads to more compact models with less accuracy drop.}
\label{tab:fs-cratch}
\vspace{-0.2cm}
\end{table*}



\subsection{Comparison to training a pruned model from random initialization}
In this experiment, we analyze the need of training an over-parameterized model with sparsity constraints. To this end, we compare the performance of a model trained and pruned using our VACL and pruning strategy (see Sect.~\ref{sect:method}) to the accuracy of the same architecture trained from scratch using a random initialization. We make use of the ResNet50/101 pruned models obtained in the previous experiment and compared those results (see Table~\ref{tab:imagenet}) to the accuracy achieved by training them from scratch using random initialization. In addition, we also compare our results to those obtained by training pruned models from random initialization using state-of-the-art methods~\cite{1802.00124}. Table~\ref{tab:fs-cratch} summarizes the results for this experiment. As shown, models trained from random initialization achieve a comparable accuracy to those trained using VACL and fine-tuning. The maximum absolute difference is $0.16\%$, which falls into the reasonable range of randomness among different runs. This result indicates that the proposed regularization and the train-prune pipeline could also be used to define the number of neurons in each layer of the network of a given architecture. In addition, as shown, our method achieve a significantly better trade-off between accuracy and model size when compared to state-of-the-art methods.

\begin{table}[t!]
\footnotesize
\begin{center}
\begin{tabular}{|c|c|c|} 
\hline
    Model                        & \# parameters & Test accuracy \\
    \hline
    ResNet50 from Keras          & 25.6M         & $\textbf{73.2\%}$    \\
    \hline
    VACL-GL pruned ResNet50 - Large         & 9.4M          & $72.1\%$      \\
    VACL-GL pruned ResNet50 - Small         & \textbf{6.1M}          & $70.1\%$      \\
    \hline
    ThiNet-GAP                   & 7.8M          & $70.2\%$      \\
\hline
\end{tabular}
\end{center}
\caption{ \textbf{Indoor-67 dataset for scene recognition~\cite{Quattoni_Torralba_CVPR}}. Comparisons of fine-tuning a pruned model to state-of-the-art methods.}
\label{indoor67table}
\vspace{-0.5cm}
\end{table}

\subsection{Transferring pruned models to other domains}
Finally, we evaluate whether a model trained using VACL-GL regularization generalizes to other vision tasks. To this end, we use the ResNet50 results on ImageNet, freeze its weights, and train only the final layer on the Indoor-67 dataset for scene recognition~\cite{Quattoni_Torralba_CVPR}. For training, we use the same image augmentations (horizontal flip, random zoom, translations, and channel shifts), with a learning rate of 0.04 decayed by a factor of $4$ every 10 epochs, over $40$ epochs. We compare our results to those obtained with a model pretrained on ImageNet whose final layer was fine-tuned on the Indoor-67 dataset, and to the variants proposed in~\cite{Luo_2017_ICCV}. Table \ref{indoor67table} shows the summary of these results. 
As shown, the model trained with the proposed VACL-GL regularization compares favorably to existing approaches both in terms of model size and test accuracy.

\section{Conclusion}
In this paper, we propose Variance-Aware Cross-Layer regularization (VACL) for pruning deep networks that have skip-connections and element-wise operations. Our approach first extends the grouping of neurons across layers to enforce aligned sparsity, and then takes into account both the first and second order statistics to constrain the variance of weights within a group. As a result we obtain a significant improvement compared to other sparsity promoting regularization techniques. We demonstrate the effectiveness of our approach for training ResNet and ResNeXt models on CIFAR, and ImageNet. Our experimental results out-perform other state-of-the-art pruning methods. In addition, our experiments demonstrate that models trained using the proposed VACL also generalize well and can be used in a transfer learning setting for Indoor scene recognition.


{\small
\bibliographystyle{ieee}
\bibliography{egbib}
}

\end{document}